\documentclass[twoside,twocolumn]{article}

\usepackage{blindtext} 

\usepackage[sc]{mathpazo} 
\usepackage[T1]{fontenc} 
\usepackage{microtype} 

\usepackage[english]{babel} 

\usepackage[hmarginratio=1:1,top=32mm,columnsep=20pt]{geometry} 
\usepackage[hang, small,labelfont=bf,up,textfont=it,up]{caption} 
\usepackage{booktabs} 

\usepackage{lettrine} 

\usepackage{enumitem} 
\setlist[itemize]{noitemsep} 

\usepackage{abstract} 

\usepackage{titlesec} 
\renewcommand\thesection{\Roman{section}} 
\renewcommand\thesubsection{\roman{subsection}} 
\titleformat{\section}[block]{\large\scshape\centering}{\thesection.}{1em}{} 
\titleformat{\subsection}[block]{\large}{\thesubsection.}{1em}{} 

\usepackage{fancyhdr} 
\usepackage{titling} 

\usepackage{hyperref} 

\usepackage{graphicx}

\pretitle{\begin{center}\Huge\bfseries} 
\posttitle{\end{center}} 
\title{Joint model for intent and entity recognition} 
\author{%
\textsc{Petr Lorenc} \\
\normalsize FEE, Czech Technical University in Prague \\ 
\normalsize \href{mailto:lorenpe2@fel.cvut.cz}{lorenpe2@fel.cvut.cz} 
}
\date{\today} 
\renewcommand{\maketitlehookd}{%
\begin{abstract}
\noindent The semantic understanding of natural dialogues composes of several parts. Some of them, like intent classification and entity detection, have a crucial role in deciding the next steps in handling user input. Handling each task as an individual problem can be wasting of training resources, and also each problem can benefit from each other. This paper tackles these problems as one. Our new model, which combine intent and entity recognition into one system, is achieving better metrics in both tasks with lower training requirements than solving each task separately. We also optimize the model based on the inputs.
\end{abstract}
}

\begin{document}

\maketitle


\section{Introduction}

In recent years, a significant amount of smart speakers have been deployed and achieved great success, such as Amazon Echo, Google Home, and many others, which interact with the user through voice interactions. Natural language understanding (NLU) is critical to the performance of spoken dialogue systems. NLU typically includes many parts, typically based on the usage, but the core usually contains the intent classification and entity recognition. For example, we have a sentence 'I would like to move to London.' We want to extract the information that the intent is 'change\_address' and also recognize that the entity is 'London'. We can use this information to query a database or other knowledge resource. In some cases, for example, in chatbot system Alquist\footnote{http://alquistai.com/}, we require to catch not only the valid named entities defined here \cite{surveyNER} but also pseudo-entities. An excellent example of pseudo-entity is 'rock music' in a sentence: 'Let's listen to rock music.'


\section{Related work}

Deep learning models have been extensively explored in NLU for several years. The traditional way to deal with intent classification and named entity recognition is to encounter both tasks separately. Named entity recognition has been thoroughly studied in \cite{surveyNER}, which can be used as a source of other resources, and intent classification is the topic of \cite{surveyIntent}.

In recent years, the multi-tasks models start occurring. The promising result is shown in \cite{ernie}, where the authors present the model called ERNIE, which is trained using several tasks as knowledge masking or prediction distance between sentences. In the work \cite{att-based}, we can see a model constructed for joint slot-filling and intent classification. In general, the task of slot-filling is similar to named entity recognition. For example query: \textbf{What flights are available from pittsburgh to baltimore on thursday morning} has intent \textbf{flight info} and following slots: 

\begin{itemize}
  \item from\_city: pittsburgh
  \item to\_city: baltimore
  \item depart\_date: thursday
  \item depart\_time: morning
\end{itemize}

Nevertheless, we can approach the slots as entities. They presented results based on recurrent neural networks and softmax-based attentions mechanism. Based on that reasoning we will be using terms \textit{named entity recognition} and \textit{slot-filling} interchangeably. 

The work \cite{att-based2} also focused on slot-filling and intent classification but in comparison with \cite{att-based} are using sparse attention mechanism. The sparse constraint assigns bigger weights for important words and lower the weights or even totally ignores the less meaningful words such as “the” or “a”. 

The more recent work \cite{bert} is also focusing on slot filling. The authors were using the novel approach for getting contextualized embeddings called BERT. We will also use other types of word embeddings, namely glove and fastext, which are commonly used for dealing with text input, see \cite{glove} and \cite{fasttext}. 

To compare our results with \cite{att-based}, \cite{att-based2} and \cite{bert}, we will be working with ATIS dataset described below.

\section{Model architecture}\label{section:model}

\begin{figure}
    \centering
    \includegraphics[width=0.4\textwidth]{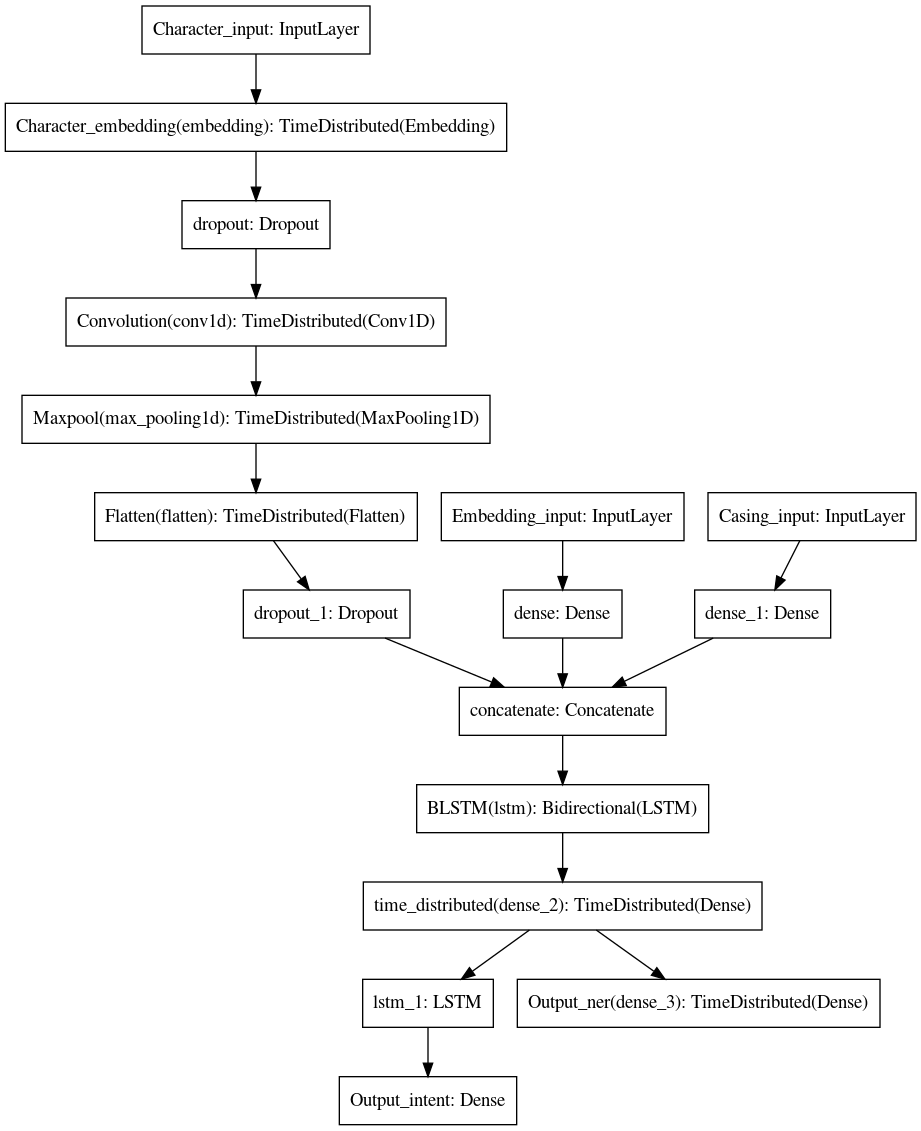}
    \caption{Model 1 of joint entity recognition and intent classification}
    \label{fig:my_model2}
\end{figure}

Based on \cite{bimodelRNN} and \cite{bert} we design a model shown in Figure \ref{fig:my_model2}. The novel approach of the model is that it uses the convolutional neural network (CNN) over char-based embeddings. It also uses features extracted from each word (such as if the word is numeric or start on lower letter) and word embedding. The architecture is word-embedding agnostic so that we can measure the metrics on different word embedding without a significant change in code. The model is also prepared for changing the architecture of the internal neural network. We have tried two types of further processing (aka types of neural networks): bidirectional Long-shot term memory\cite{lstm} (LSTM) and neural network distributed through time. The difference is that LSTM is using context over time, but in the neural network distributed through time we are using the same network at each timestep.

To find the influence of word embeddings on our result, we used three types of word representation as a vector of floats:

\begin{itemize}
  \item Glove\cite{glove}
  \item Fasttext\cite{fasttext}
  \item Bert\cite{bert}
\end{itemize}

All types of word embeddings were not fine-tuned because of possibly misleading the results (glove and fasttext cannot be fine-tuned). All files for creating embeddings are publicly available, and links are provided in github page of this project.\footnote{https://github.com/petrLorenc/poster2020}

\section{Dataset}

The dataset was created by the team in Microsoft and is publicly available\footnote{Available at github.com/Microsoft\newline/CNTK/tree/master/Examples/LanguageUnderstanding/ATIS}. The dataset (described in Table \ref{tab:atis_data}) contains spoken utterances classified into one of 26 intents. Each token in a query utterance is aligned with IOB labels. Primary, the dataset is used for intent recognition and slot filling, but as stated above, the tasks of slot filling and entity recognition are interchangeable.

    \begin{table}[h]
        \begin{center}
        \begin{tabular}{c|c}
                    \hline 
            \textbf{Name} & \textbf{\# Examples} \\\hline
            Training Set - Intent & \textit{4 978}  \\\hline
            Training Set - NER & \textit{20 426}  \\\hline
            Test Set - Intent & \textit{893}   \\\hline
            Test Set - NER & \textit{3 665}   \\\hline
        \end{tabular}
        \caption{ATIS dataset}\label{tab:atis_data}
        \label{tab}
        \end{center}
    \end{table}

\section{Results}\label{section:experiment}

    Based on model shown in \ref{section:model} we measure F1 score for NER and accuracy for intent classification on data shown in Table \ref{tab:atis_data}. These metrics were chosen because other studies are using them. The results are shown for all three types of embeddings. The results can be seen in Table \ref{tab:atis-model1-ner-embeddings} and \ref{tab:atis-model1-ir-embeddings}.

    \begin{table}[h]
        \begin{center}

            \begin{tabular}{c|c}
            \hline 
            \textbf{Embeddings} & \textbf{NER F1 accuracy}  \\\hline
            BERT & 95.78  \\\hline
            fasttext & 96.83 \\\hline
            glove & \textbf{98.44} \\\hline
            \end{tabular}
        
        \caption{Results on ATIS dataset - model 1}\label{tab:atis-model1-ner-embeddings}
        \label{tab}
        \end{center}
    \end{table}

    \begin{table}[h]
        \begin{center}

            \begin{tabular}{c|c}
            \hline 
            \textbf{Embeddings} & \textbf{Intent F1 accuracy}  \\\hline
            BERT & 92.45  \\\hline
            fasttext & 90.70 \\\hline
            glove & \textbf{95.74} \\\hline
            \end{tabular}
        
        \caption{Results on ATIS dataset - model 1}\label{tab:atis-model1-ir-embeddings}
        \label{tab}
        \end{center}
    \end{table}

\begin{figure}[t]
        \centering
        \includegraphics[width=0.4\textwidth]{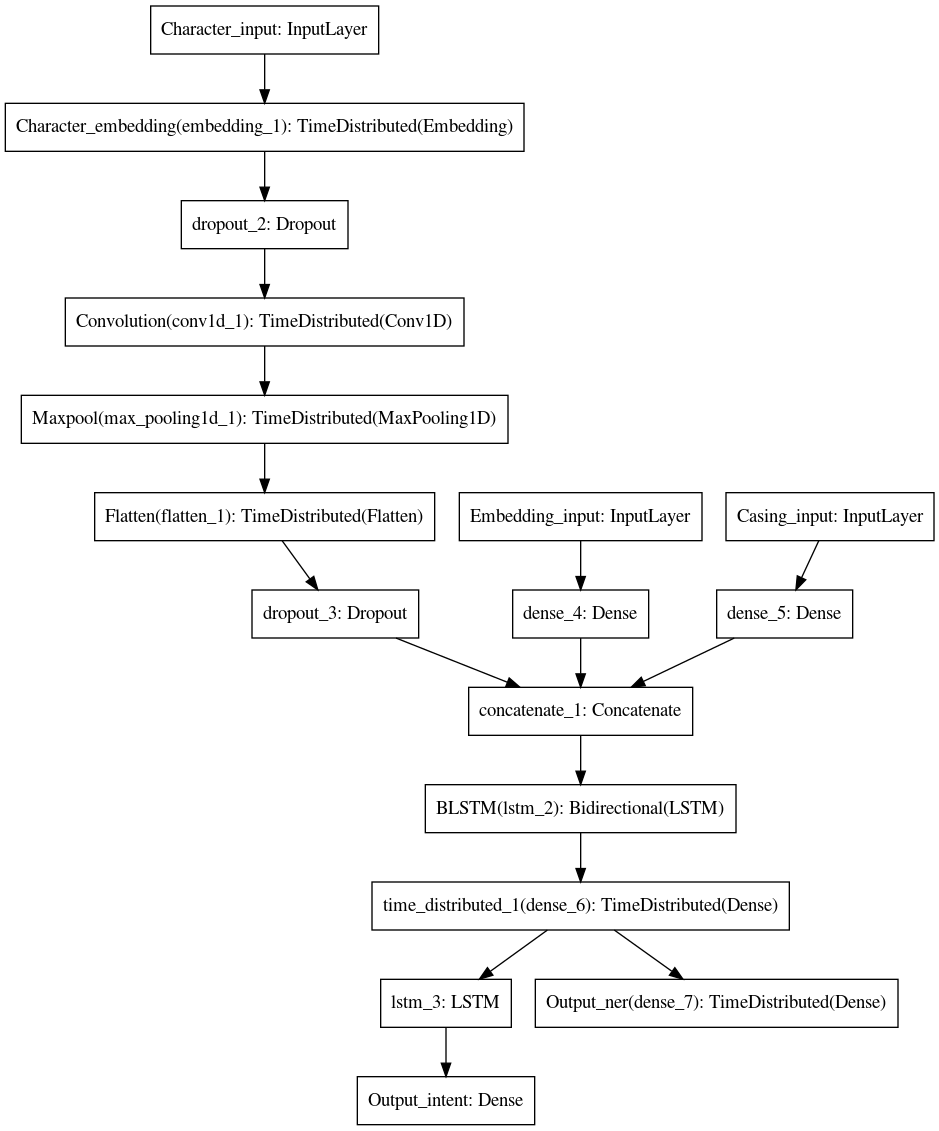}
        \caption{Model 2 of joint entity recognition and intent classification}
        \label{fig:my_model3}
    \end{figure}

\begin{figure}[b]
        \centering
        \includegraphics[width=0.4\textwidth]{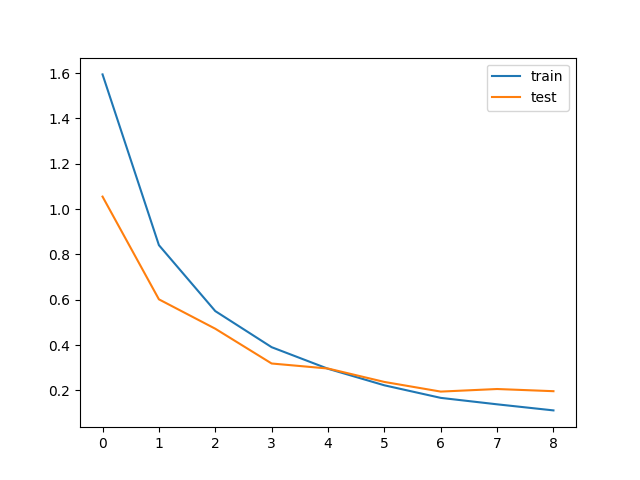}
        \caption{Train and validation loss of Glove model 1}
        \label{fig:val}
    \end{figure}

    The tables showed that Glove embeddings (the eldest approach for getting word embeddings) are getting the best results. A technique like Dropout is used in the model to avoid overfitting. Another technique was to stop training when the validation loss is not decreasing for more than two epochs. The train and validation loss are shown in Figure \ref{fig:val}. The assumption for a reason for that is because Bert should be fine-tuned, and it already has the context information in it, so the use of another context-aware neural network (like LSTM) is very useless and even adding the noise. To test that we used another network architecture, shown in Figure \ref{fig:my_model3}.

The results for the model 2 can be seen in Table \ref{tab:atis-model2-ner-embeddings} and \ref{tab:atis-model2-ir-embeddings}.

    \begin{table}[h!]
        \begin{center}

            \begin{tabular}{c|c}
            \hline 
            \textbf{Embeddings} & \textbf{NER F1 accuracy}  \\\hline
            BERT & \textbf{94.31}  \\\hline
            fasttext & 88.26 \\\hline
            glove & 88.20 \\\hline
            \end{tabular}
        
        \caption{Results on ATIS dataset - model 2}\label{tab:atis-model2-ner-embeddings}
        \label{tab}
        \end{center}
    \end{table}

    \begin{table}[h!]
        \begin{center}

            \begin{tabular}{c|c}
            \hline 
            \textbf{Embeddings} & \textbf{Intent F1 accuracy}  \\\hline
            BERT & \textbf{92.14}  \\\hline
            fasttext & 92.10 \\\hline
            glove & 89.68 \\\hline
            \end{tabular}
        
        \caption{Results on ATIS dataset - model 2}\label{tab:atis-model2-ir-embeddings}
        \label{tab}
        \end{center}
    \end{table}

    The result confirms our hypothesis that the LSTM is making classification problem harder when using BERT embeddings.
    
    The overall result is that the joint model is performing very well on these tasks with preserving or even improving the metrics on NER and intent classification tasks. For a better comparison of results, we can look at tables \ref{tab:atis-compare} and \ref{tab:atis-compare2} where we compare our algorithm with approaches from other researchers. We also include the results, which can be shown at Table \ref{tab:ir}, of the best single-task model for intent classification measured on ATIS dataset. The data for NER on this dataset is not available. In table \ref{tab:atis-compare-hw} we can see the average number of parameters of our joint models. We also measured average training time per epoch (the overall time wasn't comparable because of early stopping criteria - joint model 2 was trained for 7 epochs, the joint model 1 and only intent model was trained for 9 epochs and only ner model was trained for 12 epochs). The results of time can be seen in Table \ref{tab:atis-compare-times}. The computation was performed on 1xK80 GPU instance - ml.p2.xlarge - Amazon SageMaker ML Instance\footnote{https://aws.amazon.com/sagemaker/}.

    \begin{table}[h!]
        \begin{center}

            \begin{tabular}{c|c}
            \hline 
            \textbf{Algorithm} & \textbf{Slot/NER F1 score}   \\\hline
            Our Glove model 1 & 98.44 \\\hline
            Approach in \cite{stack-prop} & 95.90  \\\hline
            Approach in \cite{att-based} & 95.87 \\\hline
            Approach in \cite{bimodelRNN} & 96.89 \\\hline
            \end{tabular}
        
        \caption{Comparison on ATIS dataset}\label{tab:atis-compare}
        \label{tab}
        \end{center}
    \end{table}
    
    \begin{table}[h!]
        \begin{center}

            \begin{tabular}{c|c}
            \hline 
            \textbf{Algorithm} & \textbf{Intent accuracy}   \\\hline
            Our Glove model 1  & 95.74 \\\hline
            Approach in \cite{stack-prop}  & 96.90 \\\hline
            Approach in \cite{att-based}  & 98.43 \\\hline
            Approach in \cite{bimodelRNN}  & 98.99 \\\hline
            \end{tabular}
        
        \caption{Comparison on ATIS dataset 2}\label{tab:atis-compare2}
        \label{tab}
        \end{center}
    \end{table}

    \begin{table}[h!]
        \begin{center}

            \begin{tabular}{c|c}
            \hline 
            \textbf{Model\/Embeddings} & \textbf{\# parameters}   \\\hline
            Glove/Model 1  & 2,514,066 \\\hline
            Fasttext/Model 1  & 2,514,066 \\\hline
            BERT/Model 1  & 2,884,754 \\\hline
            Glove/Model 1 - Only Intent  & 2,480,656 \\\hline
            Glove/Model 1 - Only NER  & 1,982,072 \\\hline
            Glove/Model 2  & 724,882 \\\hline
            Fasttext/Model 2  & 724,882 \\\hline
            BERT/Model 2  & 910,226 \\\hline
            \end{tabular}
        
        \caption{Number of parameters}\label{tab:atis-compare-hw}
        \label{tab}
        \end{center}
    \end{table}
   
    \begin{table}[h!]
        \begin{center}

            \begin{tabular}{c|c}
            \hline 
            \textbf{Model\/Embeddings} & \textbf{Time}   \\\hline
            Glove/Model 1  & 23s \\\hline
            Glove/Model 1 - Only Intent  & 22s \\\hline
            Glove/Model 1 - Only NER  & 20s \\\hline
            Glove/Model 2  & 8s \\\hline
            \end{tabular}
        
        \caption{Training times}\label{tab:atis-compare-times}
        \label{tab}
        \end{center}
    \end{table}
    
\begin{table}[h!]
\begin{center}

    \begin{tabular}{c|c}
    \hline 
    \textbf{Algorithm} & \textbf{Intent accuracy}   \\\hline
    Approach in \cite{ir1}  & 95.66 \\\hline
    Approach in \cite{ir2}  & 95.25 \\\hline
    \end{tabular}

\caption{Intent classification task}\label{tab:ir}
\label{tab}
\end{center}
\end{table}

\subsection{Future work}

Based on \cite{att-based} and \cite{att-based2} with their models, which are using the attention neural network, we propose to explore the possibility of improving our model with the attention mechanism.

\section{Conclusion}

In this paper, we explored strategies in utilizing a joint model for named entity recognition and intent classification. The model is preserving the intent classification accuracy score and also got great results in the F1 score on NER. We compare the results of our joint model with published results in the scientific papers to get the result which we can compare. The joint model is found to be more efficient with almost the same number of parameters and requires less computational resources in comparison with separated models for each task. We also explore the importance of word embeddings on our tasks.


\section{Acknowledgements}
The research described in the paper was supervised by Ing. Jan \v{S}ediv\'y, CSc. (CIIRC CTU in Prague) and supported by the Grant Agency of the Czech Technical University in Prague, grant No. SGS20/092/OHK3/1T/37



\end{document}